\documentclass{Interspeech}

\interspeechcameraready

\usepackage{makecell}
\usepackage{multirow}
\usepackage[table]{xcolor}
\definecolor{lightgray}{gray}{0.9}

\usepackage{tipa}
\usepackage{tikz}
\usepackage{tikz-3dplot}
\usepackage{pgfplots}
\usetikzlibrary{
  shapes,
  arrows,
  positioning,
  calc,
  graphs,
  intersections,
  circuits.ee.IEC,
  quotes,
  shapes.geometric,
}

\newlength\tikzboxwidth
\newlength\tikzboxheight

\usepackage{tabularray}
\NewTblrTheme{nocaption}{
    \DefTblrTemplate{caption}{default}{}
    \DefTblrTemplate{capcont}{default}{}
}

\title{The Faetar Speech Recognition Benchmark}

\author[affiliation={1}]{Michael}{Ong}
\author[affiliation={2,3}]{Sean}{Robertson}
\author[affiliation={1,2}]{Leo}{Peckham}
\author[affiliation={1}]{Alba}{Jorquera Jimenez de Aberasturi}
\author[affiliation={1}]{Paula}{Arkhangorodsky}
\author[affiliation={1,2}]{Robin}{Huo}
\author[affiliation={1,4}]{Aman}{Sakhardande}
\author[affiliation={4}]{Mark}{Hallap}
\author[affiliation={1}]{Naomi}{Nagy}
\author[affiliation={1,2,3}]{Ewan}{Dunbar}

\affiliation{Department of Linguistics}{University of Toronto}{Canada}
\affiliation{Department of Computer Science}{University of Toronto}{Canada}
\affiliation{Department of French}{University of Toronto}{Canada}
\affiliation{Department of Philosophy}{University of Toronto}{Canada}

\email{michael.ong@mail.utoronto.ca, sdrobert@cs.toronto.edu, leo.peckham@mail.utoronto.ca,
alba.jorquera@mail.utoronto.ca,
p.arkhangorodsky@mail.utoronto.ca,
robin.huo@mail.utoronto.ca,
aman.sakhardande@mail.utoronto.ca,
mark.hallap@mail.utoronto.ca,
naomi.nagy@utoronto.ca,
ewan.dunbar@utoronto.ca}


\usepackage{comment}

\begin{document}

\maketitle

\begin{abstract}

    We introduce the Faetar Automatic Speech Recognition Benchmark, a benchmark corpus designed to push the limits of current approaches to low-resource speech recognition. Faetar, a  Franco-Proven\c cal variety spoken primarily in Italy, has no standard orthography, has virtually no existing textual or speech resources other than what is included in the benchmark, and is quite different from other forms of Franco-Proven\c cal. The corpus comes from field recordings, most of which are noisy, for which only 5~hours have matching transcriptions, and for which transcriptions are inconsistent. The corpus contains an additional 20~hours of unlabelled speech. We report baseline results from multilingual speech foundation models with a best phone error rate of 30.5\%, using a pipeline that continues pre-training on the foundation model using the unlabelled set.
\end{abstract}

\section{Introduction}

\label{sec:intro}

Speech technology for low-resource languages has been accelerated by foundation models. 
However, for the most part, demonstrations of the benefits of pre-training or cross-language transfer approaches for low-resource automatic speech recognition (ASR) have been broad-coverage evaluations aggregated over many languages and corpora at once \cite{shiFindings2023MLSUPERB2023}. This masks vast performance differences. The value of human linguistic diversity makes it important for the next era of speech recognition to deal with limited data, and speech in difficult conditions. 

We introduce a new benchmark corpus of Faetar, a variety of the Franco-Provençal language which developed in isolation in Italy, far from other speakers of Franco-Provençal,  in close contact with Italian and its dialects \cite{nagy1996language,nagy2000faetar}. Faetar has less than 1000 speakers, in Italy and in the diaspora \cite{zulato2017overview}. The benchmark represents the majority of all archived speech recordings of Faetar in existence, and is not available from any other source.

The language has no standard written form. However, as it is endangered, preservation, learning, and documentation are a priority for many community members. As such, automatic speech recognition would provide the valuable ability to transcribe and index Faetar recordings. This situation is typical of many languages, as is the quantity and quality of the data set, around five hours of labelled speech (transcribed quasi-phonetically), and around 20 hours unlabelled, from field recordings, generally noisy, of semi-spontaneous speech. 

We believe exploration and advances can be made quickly by drawing the community's attention to a common problem. The benchmark seeks to encourage researchers interested in bringing speech recognition to more languages to closely examine issues that arise in extreme conditions, rather than focusing on average-case performance, which hides a long tail with poor performance. Building a useful ASR system for Faetar is made difficult by a confluence of factors: the recordings are noisy, there are only a few hours transcribed, and, perhaps most significantly, the transcription conventions are internally inconsistent due to the lack of a standardized orthography. Most of the world's languages do not have extensive high-quality anonotated speech corpora, and many do not have standard writing systems, which means that these issues represent a major concern moving forward, yet they are likely under-represented in available corpora, which are generally biased towards making use of the best-quality data available. As such, the Faetar ASR Benchmark is an important case study.

The benchmark, including baseline and evaluation code, is freely available subject to an online data-sharing agreement.\footnote{
\url{https://perceptimatic.github.io/faetarspeech}} 
At the time of writing, the test decodings are under embargo, in order to discourage researchers from optimizing on the test set. The benchmark is thus currently available as a challenge task.


\section{Related Work}
\label{sec:related}

Low-resource speech benchmarks  began with IARPA Babel,\footnote{%
    \url{https://www.iarpa.gov/index.php/research-programs/babel}, last accessed April 8, 2024.%
} which aimed to facilitate the rapid development of robust ASR and keyword-spotting systems on languages with limited transcribed data \cite{petersonOpenASR21SecondOpen2022}. In 2020 and 2021, Babel data were repurposed for the NIST OpenASR challenges, with similar goals. 
In 2018, a low-resource speech recognition challenge for Indian languages was held \cite{srivastavaInterspeech2018Low2018}, using about 40 hours of labelled data per language.

While all of the languages in the above challenges are severely more resource-limited than English, substantial additional speech and textual data can be found for all of the languages used in the above challenges \cite{abateAmharicSpeechCorpus2005,kjartanssonCrowdsourcedSpeechCorpora2018,ardilaCommonVoiceMassivelymultilingual2020,khassanovCrowdsourcedOpensourceKazakh2021,blackCMUWildernessMultilingual2019,pratapMLSLargescaleMultilingual2020}.  Thus, in this sense, the Faetar corpus is very different from corpora used in previous challenges, as there is almost no additional publicly available data, either speech or text, that challenge participants could make use of. Furthermore, the amount of labelled data is substantially smaller than most challenges. Previous challenges also generally use higher-quality data. While other endangered languages such as Yolox\'{o}chitl Mixtec and Sierra Totonac are similarly under-resourced \cite{amithAudioCorpusYoloxochitl,amithAudioCorpusTotonac,berrebbiCombiningSpectralSelfsupervised2022,chizzoni2024,tsoukala2023asr}, these languages have not been explicitly presented to the community as benchmarks.

Recent benchmark evaluations have been developed for multilingual and low-resource speech recognition, focusing on evaluating on as many languages as possible 
\cite{javedIndicSUPERBSpeechProcessing2023,
pratapMLSLargescaleMultilingual2020,
wangVoxPopuliLargescaleMultilingual2021,
conneauFLEURSFewshotLearning2023,
shiFindings2023MLSUPERB2023}.
This breadth represents a great advance over the field's traditional attention to a few languages. However, aggregate scores  mask radical variation across languages and corpora \cite{shiFindings2023MLSUPERB2023,conneauFLEURSFewshotLearning2023,zhangGoogleUSMScaling2023,morrisOneSizeDoes2021}. Importantly, attacking the difficulties posed by the Faetar benchmark would help both with very low-resource languages without consistent orthography, and in unlocking the use of lower-quality data in higher-resource languages. This benchmark complements breadth-first approaches with the depth of a test case posing problems of significant interest.

\section{Materials} \label{sec:materials}

\subsection{Language} \label{sec:materials_language}

Faetar (pronounced [fajdar]) is spoken in Faeto, in the Apulia region of southern Italy, and is very similar to the language spoken in the neighbouring town of Celle St.~Vito. The population of both towns is less than 1000, and Faetar is threatened by population loss and language shift to Italian. The language has absorbed many Italian loan words, and, because of bilingualism, Faetar speakers often code-switch with Italian. The language is very different from Italian both in its structure and its base vocabulary. It is a variety of Franco-Proven\c{c}al, a member of the southern group of Gallo-Romance languages. Franco-Proven\c{c}al 
was brought to Apulia by a group of  
speakers between the 12th and 15th centuries (according to tradition, soldiers fighting for Charles of Anjou), and the community has since been isolated from the rest of the Franco-Proven\c{c}al speaking community \cite{nagy2000faetar,zulato2017overview}. There is also a Faetar-speaking diaspora, including in the Greater Toronto Area in Canada \cite{nagy2011lexical}.

Faetar has no standard orthography. Our recordings  were transcribed for linguistic analysis using phonetic symbols. The transcription is a mix of phonetic and phonemic transcription. For example, the word for \emph{table} is sometimes transcribed as [\textipa{tawulin@}] and sometimes  as [\textipa{tawolin:}], reflecting variability in the pronunciation; but the word for \emph{above} is always transcribed as [\textipa{indjok}] even though it is sometimes pronounced [\textipa{ijok}]. This is unlike typical spelling systems, which tend to be phonemic, but similar to transcriptions of oral collections in other languages, which often include limited dialectal variation. Given the lack of an agreed-upon orthography, it would be a large task to standardize the transcription (see Section \ref{sec:limitations}).

The phoneme inventory of Faetar is not agreed upon, but a set of core phones is given in Table \ref{tab:inventory}, based on \cite{nagy2000faetar}. In addition to the 29 phonemes in Table \ref{tab:inventory}, geminate versions of all consonants are phonemically contrastive in most cases. They are transcribed in our corpus using the length marker [{\textipa{:}}].  Some consonants have allophonic variants (/{\textipa {g}}/ can be pronounced as fricative [{\textipa {G}}]; fricatives and affricates can all be voiced), and /{\textipa {i e u o}}/ have allophonic variants [{\textipa{I E U O}}] respectively, in closed syllables. There are no phonemic length distinctions for vowels, but it is possible to have hiatus (same vowel across adjacent syllables), in which case we also used the length marker in the transcription [{\textipa{:}}], and treated the resulting doubled vowel as its own phone.  There are a number of diphthongs; in our transcription, we treat these as vowel--glide sequences. There are three additional phones transcribed that are not mentioned in \cite{nagy2000faetar}, namely [{\textipa {h}}] (which mostly occurs at the end of a word, and is rarely very audible), and both single and double [{\textipa {2}}],  a stressed variant of /{\textipa {@}}/.  In total, with geminates, doubled vowels, and variant phones, we have a 68-phone phone set.

\begin{table*}
\addtocounter{table}{-1}
\begin{talltblr}[theme=nocaption]{
  colspec={lcccccccccccccccc},
  cell{2-Z}{2-Z}=,
  vline{even[2-Z]},
  vline{1},
 cell{1}{even[2-Y]}={c=2}{c},
 hlines,
 column{2-Z}={wd=2em,c},
 columns={colsep=1pt},
 vline{odd[3-Z]}={dashed,gray6},
}
& Bilabial & & Labiodental & & Alveolar & & Postalveolar & & Palatal & & Velar  \\
Plosive & \textipa{p} & \textipa{b} & & & \textipa{t} & \textipa{d} & & & & & \textipa{k} & \textipa{g} \\
Nasal   & & \textipa{m} & & & & \textipa{n} & & & & \textltailn & & \textipa{N} \\
Trill/flap   & &  & &  & & \textipa{r} & &  \\
Fricative &  &  & \textipa{f} & \textipa{v}  & \textipa{s} &  & \textipa{S} &  &  & &  &  \\
Affricate &  &  &  &   & \textipa{ts} & & \textipa{tS} & \textipa{dZ} &   \\
Approximant & & & & & & & & & & \textipa{j} & & \textipa{w} \\
Lateral & & & & & & \textipa{l} &  & & & \textipa{L} & \\
\end{talltblr}
~~~
\quad
\begin{tikzpicture}[
  every node/.style={align=center, outer sep=0},
  node distance=0cm,
  align = center,
  dot/.style = {circle, fill, minimum size=#1,
              inner sep=0pt, outer sep=0pt},
  dot/.default = 3.5pt,
  baseline=(current bounding box.west)
]

\begin{scope}[xshift=0cm, yshift=0cm]
  \node [name=FC, dot] at (0, 0) {};
  \node [left=0pt of FC, name=FC-L] {\textipa{i}}; 
   \node[right, base right=2.3pt of FC-L, name=FC-R] {};
  \node [name=BC, dot] at (3, 0) {};
  \node [left=0pt of BC, name=BC-L] {}; \node[right, base right=2.3pt of BC-L, name=BC-R] {\textipa{u}};
  
  \node [name=FCM, dot] at (1, -1) {};
  \node [left=0pt of FCM, name=FCM-L] {\textipa{e}};
  \node[right, base right=2.3pt of FCM-L, name=FCM-R] {};
  \node [name=CCM] at (2,-1) {\textipa{@}};
  \node [name=BCM, dot] at (3, -1) {}; 
  \node [left=0pt of BCM, name=BCM-L] {}; 
  \node[right, base right=2.3pt of BCM-L, name=BCM-R] {\textipa{o}};
  
  \node [name=AA] at (2.25,-2) {\textipa{a}};
  
 \draw (FC) -- (FCM); \draw (FCM) -- (AA);
 \draw (BC) -- (BCM); \draw (BCM) -- (AA); 
 
  \draw (FC) -- (BC);
  \draw (FCM) -- (CCM); \draw (CCM) -- (BCM);

  \node at (-1.2, 0) {High};
  \node  at (-1.2, -1) {Mid};
  \node at (-1.2, -2) {Low};
  \node  at (0, 0.5) {Front};
  \node at (1.5, 0.5) {Central};
  \node  at (3, 0.5) {Back};
\end{scope}
\end{tikzpicture}

  \caption{The core phonetic inventory of Faetar. See text for discussion of additional phones used in transcriptions.
\label{tab:inventory}
  }
\end{table*}

\subsection{Data} \label{sec:materials_data}

Data were extracted from the Faetar collection of the Heritage Language Variation and Change in Toronto (HLVC) corpus \cite{nagyMultilingualCorpusExplore2011}. The corpus contains 184 recordings of native Faetar speakers collected in Faeto between 1992 and 1994 (the \emph{Homeland} subset) and 37 recordings of first- and second-generation heritage Faetar speakers collected in Toronto between 2009 and 2010  (the \emph{Heritage} subset) \cite{nagyMultilingualCorpusExplore2011}. Some of the recordings were interviews (\emph{Interview}), designed to elicit spontaneous speech, and, in others, participants were asked to describe scenes and objects from a picture book (\emph{Words}). The recordings were saved at a 44.1 kHz sampling rate (in the case of the Homeland recordings, this is the sampling rate of the digitization from analog cassette tapes). Recordings can be noisy, with considerable background noise, back-channels, and interruptions.

The source data set consisted of long-form recordings, of which 68 of the Homeland recordings, and 26 of the Heritage recordings, had been at least partially transcribed at the utterance level. Some of the transcriptions (mostly those in the Heritage subset) were extracted from ELAN annotation files, which had utterance-level alignments \cite{brugmanAnnotatingMultimediaMultimodal2004}, while the remainder were extracted from Microsoft Word files, by manually mapping numerical values from each of the two fonts used (PalPhon or IPAPhon) to modern UTF-8 encodings, and removing anything that did not look like a transcription (glosses, identifiers, \emph{etc.}). 

For the portion which was transcribed but not aligned, we adapted the JHU Arabic MGB-3 recipe for segmentation and alignment \cite{manoharJHUKaldiSystem2017}, built on Kaldi \cite{poveyKaldiSpeechRecognition2011}. This splits recordings and recording-level transcriptions by i-vector speaker diarization, splits each diarized segment into smaller, overlapping segments, aligns each overlapping segment to the text (allowing for minor changes to the text) with a seed ASR system, and  stitches the overlapping segments back together using string matching. We trained the seed system on the aligned part of the data. To generate our automatic alignments, we used a monophone, speaker-independent Gaussian mixture model-Hidden Markov model (GMM-HMM) system. We found this worked better than triphone or speaker-dependent models, presumably because of the limited aligned data available for training.  We then used PyAnnote 3.0 \cite{Plaquet23,Bredin23} to adjust utterance boundaries according to voice-activity detection and label them with speaker diarization. Because the pipeline is error-prone, we manually threw out  utterances whose boundaries were clearly misaligned or whose transcriptions were clearly wrong. On the test set (see below), we made a  more rigorous manual pass to correct alignments. 

After obtaining time-alignments for the whole data set, further filtering was performed to remove the interviewer's speech (generally clearly marked), to remove utterances with duration less than 500~ms, and to remove utterances in Italian or English.  While code-switching is common in Faetar speech (and thus, a recognizer should deal with it) removing utterances with substantial parts of Italian or English did not result in a major loss of data.  As such, we take it that the (simpler) problem we propose, which consists of transcribing Faetar with minimal code-switching, is a reasonable approximation to the real problem. For the purposes of ASR research, it will be easier to interpret the results in a more homogeneous corpus. Nevertheless, as Faetar has been in contact with Italian for a long time, and substantially longer than it has been in contact with English (English only appears in the Heritage portion of the corpus), we took a more lenient approach to identifying Italian utterances. The presence of a single English word, drawn from a closed list that were clearly not words in Faetar, was sufficient to mark an utterance as English. On the other hand, for Italian, we required that there be three words in a row, again from a closed list.


We split the aligned data into \emph{train}, \emph{dev}, and \emph{test}, ensuring a reasonable balance between male and female speakers, and between Homeland and Heritage subsets, in both \emph{train} and \emph{test}. We do not include the \emph{Words} subset in the \emph{dev} or \emph{test} splits, as we take it that this is  too easy an evaluation: many of the utterances consist of individual words, often the same word. Anticipating compatibility with the ML-SUPERB benchmark \cite{shiFindings2023MLSUPERB2023}, we also construct  \emph{1h} and \emph{10min} subsets of \emph{train}. We distribute the remainder of the data, for which we did not have transcriptions, or for which the (long-form) file could not be time-aligned, as an \emph{unlab}elled set, after obtaining VAD and speaker diarization using PyAnnote 3.0. Because we determined that many of the shorter utterances extracted in this way were not speech, we discarded all unlabelled utterances less than 1.5 seconds. Table \ref{tab:hours} shows the distribution of data in the corpus.

\begin{table}
    \centering
    \caption{
        Distribution of data in (hh:)mm:ss.
    } \label{tab:hours}
    \small
    \begin{tabular}{rrcc>{\centering\arraybackslash}p{0.5cm}>{\centering\arraybackslash}p{0.5cm}>{\centering\arraybackslash}p{0.9cm}}
        \toprule
                                &       & \multicolumn{2}{c}{Homeland} & \multicolumn{2}{c}{Heritage} & \multirow{2}{*}{Total} \\
                                &       & M     & F     & M     & F    & \\
        \midrule
        dev                     & Int.    & 8:43 & 0:31 & 0:00 & 2:35 & 11:49 \\
        \midrule
        test                    & Int.    & 13:10 & 13:33 & 10:54 & 9:16 & 46:54 \\
        \midrule
        \multirow{2}{*}{train}  & Words    & 42:31 & 58:29 & 51:44 & 9:07 & \multirow{2}{*}{4:30:17} \\
                                & Int.    & 15:58 & 27:56 & 32:49 & 14:41 & \\
        \midrule
        1h                      & Int.    & 11:08 & 13:44 & 32:50 & 0:51 & 58:34 \\
        \midrule
        10min                     & Int.    & 0:00 & 4:25 & 5:00 & 0:24 & 9:49 \\
        \midrule
        \multirow{3}{*}{unlab}    & Words    & \multicolumn{2}{c}{7:29:11} & \multicolumn{2}{c}{0:05:24}& \multirow{3}{*}{19:55:21} \\
            & Int.    & \multicolumn{2}{c}{8:23:18} & \multicolumn{2}{c}{0:51:31}&  \\
            & Mixed    & \multicolumn{2}{c}{2:12:47} & \multicolumn{2}{c}{0:53:09}& \\
        \bottomrule
    \end{tabular}
\end{table}

\section{Evaluation} \label{sec:evaluation}

The measure of performance is phone error rate (PER) on the \emph{test} set, calculated by 
aligning reference and hypothesis transcripts by the unit-cost Levenshtein algorithm. Given all utterances $u$ in a partition $\mathcal{U}$, the number of phones in a reference transcript $n_u$, and the number of errors in a model's hypothesis transcript $e_u$, the PER on $\mathcal{U}$ is defined as:
\begin{equation}
    \text{PER}_{\mathcal{U}} = \frac{\sum_{u \in \mathcal{U}} e_u}{\sum_{u \in \mathcal{U}} n_u} \times 100\%. \label{eq:per}
\end{equation}
The phonetic transcriptions contain spaces between words, making both word error rate (WER) and character error rate (CER) possible alternative metrics.  WER is too severe as the transcriptions include substantial phonetic variation that would not ordinarily be present in orthographic transcripts.  PER and CER (which differ by the exclusion or inclusion of spaces, and by the treatment of affricates and long phones) do not differ by more than a few percentage points in our observations. We chose PER (no spaces, affricates and long phones single units) to avoid making the task unnecessarily difficult. We return to PER in the context of inconsistent transcriptions in Section \ref{sec:limitations}.

\section{Baseline Experiments} \label{sec:experiments}

 We consider \emph{unconstrained} and \emph{constrained} baseline systems, where by ``constrained'' we mean that the  model is not trained on data sets other than \emph{train} (thus precluding pre-trained models trained on other languages, as well as \emph{unlab}). 
Considering constrained approaches separately allows us to focus on questions of model architecture, rather than on pre-training. 
For constrained models, as discussed in Section \ref{sec:materials_data}, force-alignment of transcriptions to audio was performed with Kaldi \cite{poveyKaldiSpeechRecognition2011}, initially using a monophone, speaker-independent HMM-GMM model, then refined at the end using a triphone, speaker-dependent model; both were composed with a 5-gram, character-level, modified Kneser-Ney language model \cite{chenEvaluationMetricsLanguage2008} trained on the (post-alignment) transcriptions from the \emph{train} partition. We report performance in Table \ref{tab:experiments}, $\pm$ the half width of a 95\% bootstrap confidence interval calculated using $K=10000$ samples. 

\begin{table}

    \centering
    \caption{%
        Phone error rates (PERs, \%) for each model (rows) and partition (columns), $\pm$  $\frac{1}{2}$-width of 95\%  confidence interval.
\label{tab:experiments}
    }%
     \rowcolors{2}{}{lightgray}
     \begin{tabular}{>{\raggedleft\arraybackslash}p{1.865cm}|>{\centering\arraybackslash}p{0.7cm}
     >{\centering\arraybackslash}p{1cm}
     >{\centering\arraybackslash}p{0.6cm}
     >{\centering\arraybackslash}p{0.5cm}
     >{\centering\arraybackslash}p{0.6cm}}
                      & Training  & Type & Train & Dev & Test  \\
     \Xhline{3\arrayrulewidth}
     Mono +5gr   & train      & Constr    & 47.1 $\pm 0.6$   & 65.9 $\pm 1.8$ & 62.6 $\pm 0.8$ \\
    Tri +5gr   & train    & Constr& 9.6 $\pm 0.4$ & 58.2 $\pm2.0$ & \textbf{56.7} $\pm 0.9$ \\
     \Xhline{3\arrayrulewidth}
     MMS Fine-tune (FT)      & train    &      & 34.4 $\pm 0.5$& 39.9 $\pm 2.8$   & \textbf{33.0} $\pm 0.8$ \\
     mHuBERT-147 FT      & train    &      & 26.6 $\pm 0.5$ & 41.1  $\pm 2.6$  & 33.6 $\pm 0.8$ \\
     ESPnet-MMS    & train  & &   36.0 $\pm 0.5$ & 43.0 $\pm 2.8$ & {35.9} $\pm 0.8$ \\
    \Xhline{3\arrayrulewidth}
    MMS Pre-train (PT)   & train  & + unlab &   31.4 $\pm 0.5$    & 38.7  $\pm 2.7$   &   31.5 $\pm 0.8$    \\
    MMS Self-train (ST)   & train  & + unlab &   31.4 $\pm 0.5$  & 36.6  $\pm 2.7$   &   31.0 $\pm 0.8$   \\
    MMS PT + ST   & train  & + unlab &   30.0 $\pm 0.5$   & 36.4  $\pm 2.7$  &   \textbf{30.5} $\pm 0.8$  \\
\Xhline{3\arrayrulewidth}
    ESPnet-MMS  & 1hr    & &   39.3 $\pm 0.5$ & 44.4 $\pm 2.8$  & 37.4 $\pm 0.8$ \\
    ESPnet-MMS & 10min    & &  48.2 $\pm 0.5$  & 50.2 $\pm 2.4$  & 45.1 $\pm 0.8$ \\    
    \end{tabular}
\end{table}

In the \emph{unconstrained} condition, we fine-tune existing multilingual models MMS \cite{pratapScalingSpeechTechnology2023} and mHuBERT-147 \cite{boito2024mhubert} on the \emph{train} set. Both are masked prediction models; MMS is a wav2vec 2.0 model that takes a language-specific adapter, head, and fine-tuning strategy. Rather than initializing the adapter layers from scratch, we initialize the adapters and head with a model fine-tuned on Italian.
mHuBERT-147 takes an iterative refinement approach, training first on classification according to classes derived from k-means clustering on spectral features, then, after the first iteration, on targets derived from clustering the internal representations of the previous one. As \cite{boito2024mhubert} obtained best results with three iterations, we fine-tune the third iteration. For both models, we use a linear schedule with peak learning rate of $1e^{-5}$ for 200 epochs of which ten are warmup, dropout of 10\%, and effective batch size of 8. 
For reference, we also train using the  ML-SUPERB recipe \cite{shiFindings2023MLSUPERB2023} from ESPNet 
\cite{watanabeESPnetEndtoendSpeech2018}, which also fine-tunes the 1-billion parameter MMS model, keeping all hyperparameters as-is except to set the effective batch size to 4. For comparison with ML-SUPERB, we also include results for the restricted \emph{1hr} and \emph{10min} training conditions. 

The  data includes an \emph{unlab}elled subset. We show two possible approaches to using it. First, \emph{continued pre-training} of a multilingual model using Faetar audio for self-supervised training prior to fine-tuning. We use the entire \emph{unlab} set and continue pre-training of MMS (effective batch size 32; 300000 steps $\approx$258 epochs, of which the first 30\% are warmup in a linear scheduler; peak learning rate $3\times10^{-4}$), then fine-tune as before.  As can be seen from Table \ref{tab:experiments}, this leads to a large performance boost. 
Alternatively, we attempt \emph{self-training,} decoding \emph{unlab} using the fine-tuned (base) MMS model, then adding the resulting decoded speech to a second round of fine-tuning.  Finally, we combine both approaches. Table \ref{tab:experiments} demonstrates that both of these approaches improve performance, and that the results are cumulative, leading to a best PER of 30.5\% when combining both approaches (although the pairwise differences are within the overlap of the confidence intervals again). In sum, the unlabelled data set shows potential for improving ASR.

\section{Limitations} \label{sec:limitations}

The major limitation of this benchmark is that the transcriptions themselves are inconsistent. As discussed above, they vary unpredictably between tracking fine-grained phonetic distinctions and tracking lexical-level (phonemic) distinctions; as with many collections of linguistic data, these were collected over many years, and transcribed for various different purposes. Furthermore, as is often the case in field linguistics when fine-grained phonetic details are being transcribed (particularly from unclear recordings) and there is no clear consensus about what the phonemic form is, the transcriptions are inconsistent in another sense, namely that specific phonetic details may not always be transcribed in exactly the same way. Genuine errors were corrected in the test set, but these two kinds of inconsistency remain, even in the test set. These two sources of noise in the transcription mean that there is an intrinsic lower bound (``noise threshold'') for the phone error rate which is unknown.

The fact that we are limited to transcriptions of this kind is partly a consequence of the fact that Faetar has no standard orthography and that the linguistic (phonemic) analysis is not fully understood. The fact that there is no standard orthography does not detract from the utility of ASR: a major application of ASR for minoritized and endangered languages is in indexing audio recordings and in generating textual resources from them (that may, for example, later be used to build educational materials), in addition to linguistic research. Document collections can be built without a standard orthography.

However, for most purposes---even for linguistic phonetic research---having fine-grained phonetic transcriptions (even perfectly consistent ones) is insufficient. For most practical purposes, it is critical to have a lexical transcription which indicates the intended word but does not encode all details of its pronunciation---for example, grouping [\textipa{tawulin@}] and  [\textipa{tawolin:}] together. The serious challenge posed here, and in other similar corpora, is that, even armed with an extensive dictionary of Faetar, we would have no way of automatically mapping our sub-lexical transcriptions to the headwords in the dictionary, and, more importantly, of knowing whether we have accurately done so. More importantly, existing lexica are highly incomplete. Previous approaches to inconsistent transcriptions generally rely on extensive manual labelling or elicitation of alternate transcriptions of the same utterance \cite{ali2015multi,nigmatulina2020asr,barends2024automatic}. Both would be prohibitively time- and expert-intensive for us.

We take the future challenge of guessing at an ASR lexicon or a phonemic-level transcription convention based in a fully unsupervised manner, with no lexical-level grouping or annotation used in training, to be potentially ground-breaking for low-resource languages. We do not explore this except to note that informal experiments with language modelling based on lexica derived by clustering  did not yield improvements to PER.

\section{Summary of Contributions} \label{sec:conclusions}

We have presented the Faetar Automatic Speech Recognition Benchmark, a benchmark corpus aimed at under-resourced languages with noisy audio and transcriptions. We demonstrate that out-of-the-box application of state-of-the-art multilingual foundation models yields an improvement over spectral features, and that using unlabelled data yields further improvements. The fact that we obtain improvements on the constrained ASR task by varying models suggests that there may be further improvements to be made to the  model architectures as well. Nevertheless, in all cases, the baseline PERs would generally be unacceptable without manual correction. We leave additional approaches (e.g., improved speaker modelling, speech enhancement, hybrid models) for the community to explore, and  release the benchmark with the aim of encouraging careful attention to corpora and languages like Faetar, deepening the field's commitment to the  promise of universal speech technology.

\bibliographystyle{IEEEtran}
\bibliography{mybib}

\end{document}